\documentclass{article}

\usepackage{microtype}
\usepackage{graphicx}
\usepackage{subfigure}
\usepackage{booktabs} %

\usepackage{hyperref}

\usepackage[accepted]{icml2024}
\usepackage{fancyhdr}

\usepackage{amsmath}
\usepackage{amssymb}
\usepackage{mathtools}
\usepackage{amsthm}

\usepackage{natbib}
\usepackage[capitalize,noabbrev]{cleveref}

\usepackage[algo2e,linesnumbered,noend]{algorithm2e}
\SetKwInput{KwEn}{Encoding}
\SetKwInput{KwDe}{Decoding}

\theoremstyle{plain}
\newtheorem{theorem}{Theorem}[section]
\newtheorem{proposition}[theorem]{Proposition}
\newtheorem{lemma}[theorem]{Lemma}
\newtheorem{corollary}[theorem]{Corollary}
\theoremstyle{definition}

\theoremstyle{remark}
\newtheorem{remark}[theorem]{Remark}

\usepackage{xcolor}
\usepackage[showdeletions]{color-edits}	
\usepackage{booktabs}
\usepackage{braket}
\usepackage{multicol}
\usepackage{multirow}
\usepackage{array}

\usepackage[textsize=tiny]{todonotes}

\usepackage{stfloats}

\DeclareMathOperator*{\argmax}{arg\,max} 
\DeclareMathOperator{\diag}{diag}

\newcolumntype{C}[1]{>{\centering\arraybackslash}m{#1}}

\def\reals{\mathbb{R}}

\def\Bc{{\cal B}}

\def\nn{\nonumber}
\def\beq{\begin{equation}}
\def\eeq{\end{equation}}
\def\beqa{\begin{eqnarray}}
\def\eeqa{\end{eqnarray}}
\def\balign{\begin{align}}
\def\ealign{\end{align}}
\def\bpr{\begin{proof}}
\def\epr{\end{proof}}
\def\bth{\begin{theorem}}
\def\eth{\end{theorem}}
\def\blm{\begin{lemma}}
\def\elm{\end{lemma}}
\def\bprop{\begin{proposition}}
\def\eprop{\end{proposition}}
\def\bcr{\begin{corollary}}
\def\ecr{\end{corollary}}

\def\eg{{\it e.g.,\ \/}}

\def\and {{\rm and}}

\icmltitlerunning{Improving SAM Requires Rethinking its Optimization Formulation}

\begin{document}

\twocolumn[
\icmltitle{Improving SAM Requires Rethinking its Optimization Formulation}

\icmlsetsymbol{equal}{*}

\begin{icmlauthorlist}
\icmlauthor{Wanyun Xie}{lions}
\icmlauthor{Fabian Latorre}{lions}
\icmlauthor{Kimon Antonakopoulos}{lions}
\icmlauthor{Thomas Pethick}{lions}
\icmlauthor{Volkan Cevher}{lions}
\end{icmlauthorlist}

\icmlaffiliation{lions}{Laboratory for Information and Inference Systems, École Polytechnique Fédérale de Lausanne (EPFL), Switzerland}

\icmlcorrespondingauthor{Wanyun Xie}{wanyun.xie@epfl.ch}

\icmlkeywords{Machine Learning, ICML}

\vskip 0.3in
]

\printAffiliationsAndNotice{}  %

\begin{abstract}

This paper rethinks Sharpness-Aware Minimization (SAM), which is originally formulated as a zero-sum game where the weights of a network and a bounded perturbation try to minimize/maximize, respectively, the same differentiable loss.
To fundamentally improve this design, we argue that SAM should instead be reformulated using the 0-1 loss. 
As a continuous relaxation, we follow the simple conventional approach where the minimizing (maximizing) player uses an upper bound (lower bound) surrogate to the 0-1 loss. 
This leads to a novel formulation of SAM as a bilevel optimization problem, dubbed as BiSAM. 
BiSAM with newly designed lower-bound surrogate loss indeed constructs stronger perturbation.
Through numerical evidence, we show that BiSAM consistently results in improved performance when compared to the original SAM and variants, while enjoying similar computational complexity.
Our code is available at \url{https://github.com/LIONS-EPFL/BiSAM}.

\end{abstract}

\section{Introduction}\label{sec:intro}
The rise in popularity of deep neural networks has motivated the question of which optimization methods are better suited for their training. Recently, it has been found that \textit{Sharpness-Aware Minimization (SAM)} \citep{sam20} can greatly improve their generalization with almost negligible increase in computational complexity. 
SAM not only benefits supervised learning tasks from computer vision greatly \citep{sam20,dosovitskiy2021an} but also improves the performance of large language models \citep{Bahri2021SharpnessAwareMI, sam-language22}. Hence, it is natural to ask whether SAM can be improved further. Indeed, many works have been quick to present modifications of the original SAM algorithm, that improve its speed \citep{du2022efficient} or performance \citep{asam21} in practice.

The motivation behind SAM is to find a parameter $w^\star$ in the so-called \textit{loss landscape}, that achieves a low \textit{loss value} while being \textit{flat} i.e., the loss in its immediate neighborhood should not deviate meaningfully from the value attained at $w^\star$. 
To seek the flat minima, SAM algorithm derived by \citet{sam20} is formulated as a two-player zero-sum game, where two players respectively seek to minimize and maximize the same differential loss.

In supervised classification tasks, common differential losses like cross-entropy provide an upper bound to misclassification error, making it reasonable to expect that minimizing these losses will lead to a reduction in the 0-1 loss for the minimizer.
However, can we do such a natural replacement in the SAM formulation? A key question is \textit{``Does maximizing such surrogate loss lead to a maximum of the 0-1 loss?"} 
No, there is no guarantee. We make this precise with a counterexample in \cref{subsec:SAM_limitation}.

Therefore, we argue that even though the surrogate loss used in SAM formulation appears beneficial, we should recall that the goal in supervised classification is not to achieve a low value of the cross-entropy, rather, the goal is to enjoy a small \textit{misclassification error rate} on the testing set. This raises the question:
\begin{center}
\textit{If our goal is to achieve a better classifier, should we not apply the SAM formulation directly on the misclassification error i.e., the 0-1 loss?}
\end{center}

Motivated by this actual goal, we argue that the original SAM with surrogate loss suffers from a fundamental flaw that leads to a weaker perturbation.
Our analysis reveals that maximizing a surrogate loss like cross-entropy has no guarantee that misclassification errors will increase due to its upper-bound nature. To overcome this shortcoming, the minimizer and maximizer should have different objectives, specifically an upper bound of misclassification error for the minimizer while a lower bound for the maximizer. 
This guides us to propose a novel bilevel formulation of SAM, called BiSAM, with a new loss function for the maximizer. BiSAM fixes SAM's fundamental issue without any extra computational cost and importantly it can be incorporated in almost all existing variants of SAM.

We summarize \textbf{our contributions} as follows:
\begin{itemize}
\item We present a novel bilevel optimization formulation of SAM, where instead of solving a min-max zero-sum game between the model parameters $w$ and the perturbation $\epsilon$, each player has a different objective. This formulation appears naturally by applying SAM to the relevant performance metric in supervised learning: the misclassification error, i.e., the so-called 0-1 loss. 

\item We propose \textit{BiSAM} (\cref{alg:BiSAM}), a scalable first-order optimization method to solve our proposed bilevel formulation of SAM. BiSAM is simple to implement and enjoys a similar computational complexity when compared to SAM.
 
\item We present numerical evidence on CIFAR10 and CIFAR100 showing that BiSAM consistently outperforms SAM across five models, and also see improvement on ImageNet-1K. BiSAM incorporating variants of SAM (ASAM and ESAM) also demonstrates enhancement. We additionally verify that our reformulation remains robust in fine-tuning both image and text classification tasks and noisy label tasks.
\end{itemize}

\section{Preliminaries and Problem Setup}\label{sec:background}

\textbf{Notation.} 
$f_w: \mathbb{R}^d \to \mathbb{R}^K$ corresponds to the logits (scores) that are output by a neural network with parameters $w$. For a given \textit{loss function} $\ell$, we denote the \textit{training set loss} $L(w) = \frac{1}{n} \sum_{i=1}^n \ell(f_w(x_i), y_i)$. We denote the corresponding training set losses of cross entropy loss and 0-1 loss as $L^\text{ce}$ and $L^{01}$, respectively. 
We denote as $\mathbb{I}\{A\}$ the indicator function of an event i.e., $\mathbb{I}\{A\} = 1$ if $A$ is true or $\mathbb{I}\{A\}=0$ if $A$ is false.

\subsection{Preliminaries: Surrogate loss}
Considering a classification task, the goal is to obtain a classifier predicting the label $y$ correctly from data $x$. Then, the problem is
\begin{equation}
    \min_w L^{01} (w) = \min_w \frac{1}{n} \sum_{i=1}^n \mathbb{I} \left \{ \argmax_{j=1,\ldots,K} f_{w}(x)_j \neq y \right \} 
\label{eq:minimum01}
\end{equation}
The difficulty of solving \cref{eq:minimum01} is the fact that $L^{01}$ is discontinuous, which hinders first-order optimization methods. Normally, it can be solved by replacing $L^{01}$ by a surrogate loss. For minimization problems, a differential upper bound of misclassification error can be used, i.e. 
\begin{equation}
    L^{01} (w) \leq L (w)
\label{eq:upperbound}
\end{equation}
Common surrogate losses are cross-entropy loss and hinge loss. It is important to note that this inequality \cref{eq:upperbound} guarantees that minimizing $L(w)$ provides a solution that decreases $L^{01}$, which is the real goal in supervised classification tasks.

\subsection{SAM and its limitation}
\label{subsec:SAM_limitation}
SAM \citep{sam20} and its variants \citep{asam21,du2022efficient} search for a flat minima by solving the following minimax optimization problem:
\begin{equation}
    \min_w \max_{\epsilon:\| \epsilon\|_2\leq\rho} L (w+\epsilon)
\label{eq:sam}
\end{equation}
where $\rho$ is a small constant controlling the radius of a neighborhood, and a surrogate upper-bound loss is usually used as the objective function following the minimization problem. 
SAM addresses this minimax problem by applying a two-step procedure at iteration $t$:
\begin{equation}
\begin{split}
    &\epsilon_t =  \rho \frac{\nabla L(w_t)}{\| \nabla L(w_t) \|_2} \approx \argmax_{\epsilon:\| \epsilon\|_2\leq\rho} L(w_t+\epsilon)
     \\ 
    &w_{t+1} = w_t - \eta_t \nabla L(w_t+\epsilon_t) %
\end{split}
\label{eq:sam2}
\end{equation}
where the perturbation $\epsilon_t$ is obtained as the closed-form solution to a first-order approximation. 

Although this approach is pervasive in practice, we argue that the surrogate loss for the inner maximization problem has a fundamental limit that will lead to \textbf{weaker perturbation}. To be specific, the maximizer in \cref{eq:sam2} maximizes an \textit{upper bound} on the classification error. This means that any $\epsilon^\star$ obtained by \cref{eq:sam2} has no guarantee to increase the classification error.
To make our argument more convincing, we give an example.

\paragraph{Example.}
This example is to illustrate explicitly that maximizing the upper bound of 0-1 loss leads to the \textit{wrong} result. Consider two possible vectors of logits:
\begin{itemize}
    \item ``Option A'': $(1/K + \delta, 1/K - \delta, \ldots, 1/K)$
    \item ``Option B'': $(0.5 - \delta, 0.5 + \delta, 0, 0, \ldots, 0)$
\end{itemize}
where $\delta$ is a small positive number, and assume the first class is the correct one, so we compute the cross-entropy with respect to the vector $(1, 0, \ldots, 0)$.

For ``Option A'', the cross-entropy is $-\log(1/K + \delta)$, which tends to infinity as $K\to \infty$ and $\delta \to 0$.  For ``Option B'', the cross-entropy is $-\log(0.5-\delta)$, which is a small constant number. Hence, an adversary that maximizes an upper bound like the cross-entropy, would always choose ``Option A''. However, note that ``Option A'' \textit{never} leads to a maximizer of the 0-1 loss, since the predicted class is the correct one (zero loss). In contrast, ``Option B'' always achieves the maximum of the 0-1 loss (loss is equal to one), even if it has low (i.e., constant) cross-entropy. This illustrates why maximizing an upper bound like cross-entropy provides a possibly weak weight perturbation.

The misalignment observed above suggests that we should take one step back and rethink SAM on $L^{01}$ directly, as it directly reflects the accuracy of interest:
\begin{equation}
    \min_w \max_{\epsilon:\| \epsilon\|_2\leq\rho} L^{01} (w+\epsilon)
\label{eq:sam01}
\end{equation}
Handling the discontinuous nature of $L^{01}$ to fit the first-order optimization methods, we decouple \cref{eq:sam01} into a bilevel formulation and treat the maximizer and the minimizer separately in \cref{sec:bisam}.

\section{Bilevel Sharpness-aware Minimization (BiSAM)}\label{sec:bisam}
In order to obtain a differentiable objectives for the minimization and maximization players in the formulation
\cref{eq:sam01}, our starting point is to decouple the problem as follows:
\begin{equation}
\begin{aligned}
    &\min_{w} L^{01}(w + \epsilon^\star), \\
    &\text{subject to } \epsilon^\star \in \argmax_{\epsilon: \|\epsilon\|_2 \leq\rho} L^{01}(w + \epsilon)
\end{aligned}
\label{eq:bilevel}
\end{equation}
Up to this point, there has been no modification of the original objective \cref{eq:sam01}. We first note that for the minimization player $w$, we can minimize a differentiable upper bound (e.g., cross-entropy) instead of the 0-1 loss, leading to the formulation:
\begin{equation}
\begin{aligned}
    &\min_{w} L^{\text{ce}}(w + \epsilon^\star), \\
    &\text{subject to } \epsilon^\star \in \argmax_{\epsilon: \|\epsilon\|_2 \leq\rho} L^{01}(w + \epsilon)
\end{aligned}
\label{eq:bilevel2}
\end{equation}
Now, we only need to deal with replacing the 0-1 loss in the objective of the perturbation $\epsilon$. Because this corresponds to a maximization problem, we need to derive a \textit{lower bound}.

\paragraph{Recall the example.}
We maximize a lower bound instead of the cross-entropy in the example in \cref{subsec:SAM_limitation}. Consider a lower-bound loss like $\max_{j \in [K]} \tanh( z_j - z_y)$ where $z$ is the vector of logits (see \eqref{eq:lem_proof_eq1} regarding the construction). This loss for ``option A'' is $0$ and for ``option B'' it is $\tanh(2\delta)>0$. Thus, an adversary maximizing this loss would choose ``option B'', correctly leading to maximization of the 0-1 loss.

\begin{lemma}\label{lem:lower_bound_epsilon}
Let $\phi(x)$ be a lower bound of the 0-1 step function $\mathbb{I}\{x > 0\}$.
For each $j \in [K]$, let $F_{w+\epsilon}(x_i, y_i)_j = f_{w+\epsilon}(x_i)_j - f_{w+\epsilon}(x_i)_{y_i}$ and let $\mu > 0$. It holds that
\begin{equation}
\begin{split}
    L^{01}(w+\epsilon) \geq 
    & \frac{1}{n} \sum_{i=1}^n \frac{1}{\mu} \log \left(\sum_{j=1}^K e^{\mu \phi({F_{w + \epsilon} (x_i,y_i)_j})} \right) \\ &- \frac{1}{\mu} \log(K)
\end{split}
\end{equation}
\end{lemma}
\begin{remark}
Choice $\phi(x)=\tanh(x)$ is a valid lower bound (see \eqref{eq:tanh} for further discussion).
\end{remark}

\begin{proof}
Note that for a training sample $(x_i, y_i)$ we have misclassification error if and only if for some class $j \neq y_i$ the score assigned to class $j$ is larger than the score assigned to $y_i$. Equivalently,
$\argmax_{j \in [K]} f_{w + \epsilon}(x_i)_j \neq y_i$ if and only if $\max_{j=1, \ldots K} F_{w+\epsilon}(x_i, y_i)_j > 0$. Thus,
\begin{equation}\label{eq:lem_proof_eq1}
\begin{split}
L^{01}(w+\epsilon) &= \frac{1}{n} \sum_{i=1}^n \mathbb{I} \left \{ \argmax_{j \in [K]} f_{w + \epsilon}(x_i)_j \neq y_i \right \} \\
&= \frac{1}{n} \sum_{i=1}^n  \mathbb{I} \left\{ \max_{j\in[K]} F_{w+\epsilon}(x_i,y_i)_j > 0 \right\} \\
&= \frac{1}{n} \sum_{i=1}^n \max_{j\in[K]}  \mathbb{I} \left\{ F_{w+\epsilon}(x_i,y_i)_j > 0 \right\} \\
& \geq \frac{1}{n} \sum_{i=1}^n \max_{j\in[K]}  \phi \left ( F_{w+\epsilon}(x_i,y_i)_j\right )
\end{split}
\end{equation}
Although this lower bound can be used directly, it is computationally consuming due to the need for $K$ perturbation calculations per update.
To get rid of the non-differentiable maximum operator over the set $[K]=\{1, \ldots, K \}$, we use the well-known bounds of the log-sum-exp function:
\begin{equation}\label{eq:logsumexp_bound}
\frac{1}{\mu} \log \left ( \sum_{i=1}^K e^{\mu a_i} \right ) \leq \max \{a_1, \ldots, a_K\} + \frac{1}{\mu}\log(K)
\end{equation}
Using \cref{eq:logsumexp_bound} in \cref{eq:lem_proof_eq1} yields the desired bound.
\end{proof}

As a consequence of \cref{lem:lower_bound_epsilon} we conclude that a valid approach for the maximization player is to solve the differentiable problem in the right-hand-side of the following inequality:
\begin{equation}\label{eq:q_definition}
\begin{split}
    &\max_{\epsilon: \|\epsilon\|_2 \leq\rho} L^{01}(w + \epsilon) + \frac{1}{\mu}\log(K) \\
    &\geq  \max_{\epsilon: \|\epsilon\|_2 \leq\rho}
    \frac{1}{n} \sum_{i=1}^n \frac{1}{\mu} \log \left(\sum_{j=1}^K e^{\mu \phi({F_{w+\epsilon}(x_i,y_i)_j})} \right) 
    \\
    &\eqqcolon \max_{\epsilon: \|\epsilon\|_2 \leq\rho}  Q_{\phi, \mu}(w + \epsilon)
\end{split}
\end{equation}
Continuing from \cref{eq:bilevel2}, we finally arrive at a bilevel and fully differentiable formulation:
\begin{equation}
\begin{aligned}
    &\min_{w} L^{\text{ce}}(w + \epsilon^\star), \\
    &\text{subject to } \epsilon^\star \in \argmax_{\epsilon: \|\epsilon\|_2 \leq\rho} Q_{\phi, \mu}(w + \epsilon)
\end{aligned}
\label{eq:bilevel3}
\end{equation}
Note that both upper and lower objective functions in \cref{eq:bilevel3} are aligned with the ultimate goal of the 0-1 loss. Specifically, the bound in the lower level problem is tight up to a small additive logarithmic term of $\frac{1}{\mu} \log(K)$ and it holds everywhere.

The above setting forces us to take a slight detour to the general framework of bilevel optimization and its solution concepts. In particular, the nested nature of the problem makes its solution to be notoriously difficult. Therefore, the success of the up-to-date iterative methods relies on a set of quite restrictive assumptions, which do not apply in the complex environment of neural networks (we refer the reader to \citet{ghadimi2018approximation,tarzanagh2023online} for more details). In particular, an important feature that needs to be satisfied is that the so-called inner problem should be strongly convex; which here is clearly not the case. 
Therefore, in order to devise a fast algorithm for the problem in the right-hand-side of \eqref{eq:bilevel3}, some particular modifications should be made. 
More precisely, we follow the same approach in the original SAM algorithm \citep{sam20}, and do a first-order Taylor expansion of $Q_{\phi, \mu}(w+\epsilon)$ with respect to $\epsilon$ around 0. We obtain:
\begin{equation}
\begin{aligned}
    \epsilon^\star &\in \argmax_{\epsilon: \|\epsilon\|_2 \leq\rho} Q_{\phi, \mu}(w+\epsilon) \\
    &\approx \argmax_{\epsilon: \|\epsilon\|_2 \leq\rho} Q_{\phi, \mu}(w) + \epsilon^\top \nabla_w Q(w) \\
    &= \argmax_{\epsilon: \|\epsilon\|_2 \leq\rho} \epsilon^\top \nabla_w Q(w) 
    = \rho \frac{\nabla_w Q(w)}{\|\nabla_w Q(w)\|_2}
\end{aligned}
\label{eq:epsilon}
\end{equation}
As the function $Q_{\phi, \mu}$ involves a sum over the whole dataset, this makes the computation of the full gradient in \cref{eq:epsilon} too costly. For scalability, we use stochastic gradients defined on a mini-batch in practice. Our proposed algorithm to solve SAM in the bilevel optimization paradigm (BiSAM), finally takes shape as shown in \cref{alg:BiSAM}. 

\begin{algorithm}[!t]
\caption{Bilevel SAM (BiSAM)}
\label{alg:BiSAM}
\SetAlgoLined
\setcounter{AlgoLine}{0}
    \KwIn{Initialization $w_0 \in \reals^d$, iterations $T$, batch size $b$, step sizes $\{\eta_t\}_{t=0}^{T-1}$, neighborhood size $\rho>0$, $\mu>0$, lower bound $\phi$.}
    \For{$t=0$ {\bfseries to} $T-1$}{
        Sample minibatch $\Bc=\{(x_1, y_1), \ldots, (x_b, y_b)\}$. \par
        Compute the (stochastic) gradient of the perturbation loss $Q_{\phi, \mu}(w_t)$ defined in \cref{eq:q_definition}. \par
        Compute perturbation $\epsilon_t = \rho \frac{\nabla_w Q(w)}{\|\nabla_w Q(w)\|}$. \par
        Compute gradient $g_{t} = \nabla_w L_\Bc(w_t + \epsilon_t)$. \par
        Update weights $w_{t+1} = w_t - \eta_t g_{t}$.
    }
\end{algorithm}

\textbf{On the choice of lower bound $\phi$. } The function $\phi$ plays a crucial role in the objective $Q_{\phi, \mu}$ that defines the perturbation \cref{eq:epsilon}. Although in theory we can use any lower bound for the 0-1 step function $\mathbb{I}\{x > 0\}$, the choice can affect the performance of the optimization algorithm. As is always the case in Deep Learning, one should be on the look for possible sources of \textit{vanishing/exploding gradients} \citep{hochreiter2001gradient}.

\vspace{-1mm}
\begin{figure}[!t]
     \centering
     \subfigure[$\phi(x)=\tanh(\alpha x)$]{\includegraphics[width=0.4\textwidth]{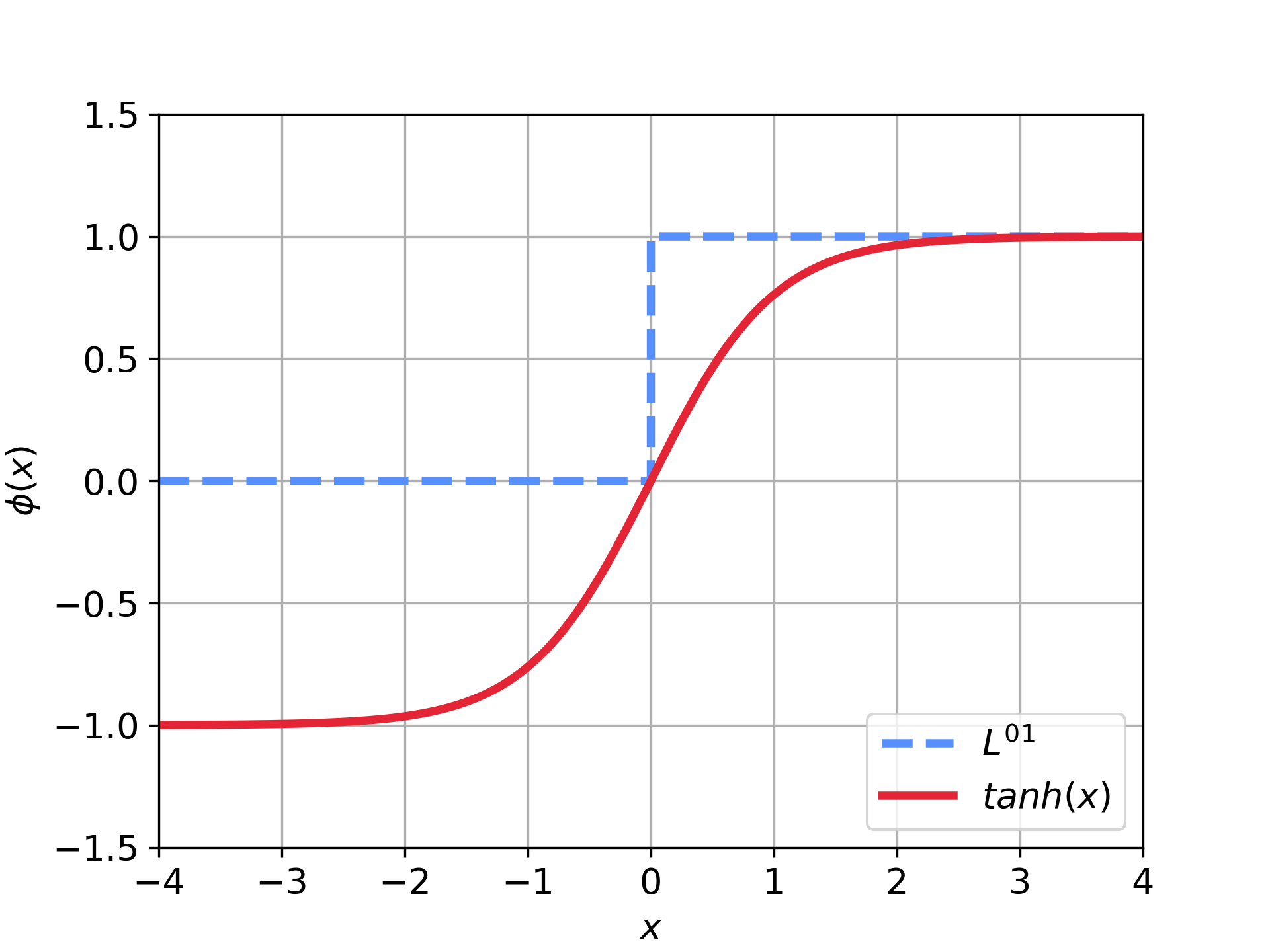}}
     \subfigure[$\phi(x)=-\log(1 + e^{(\gamma-x)}) + 1$]{\includegraphics[width=0.4\textwidth]{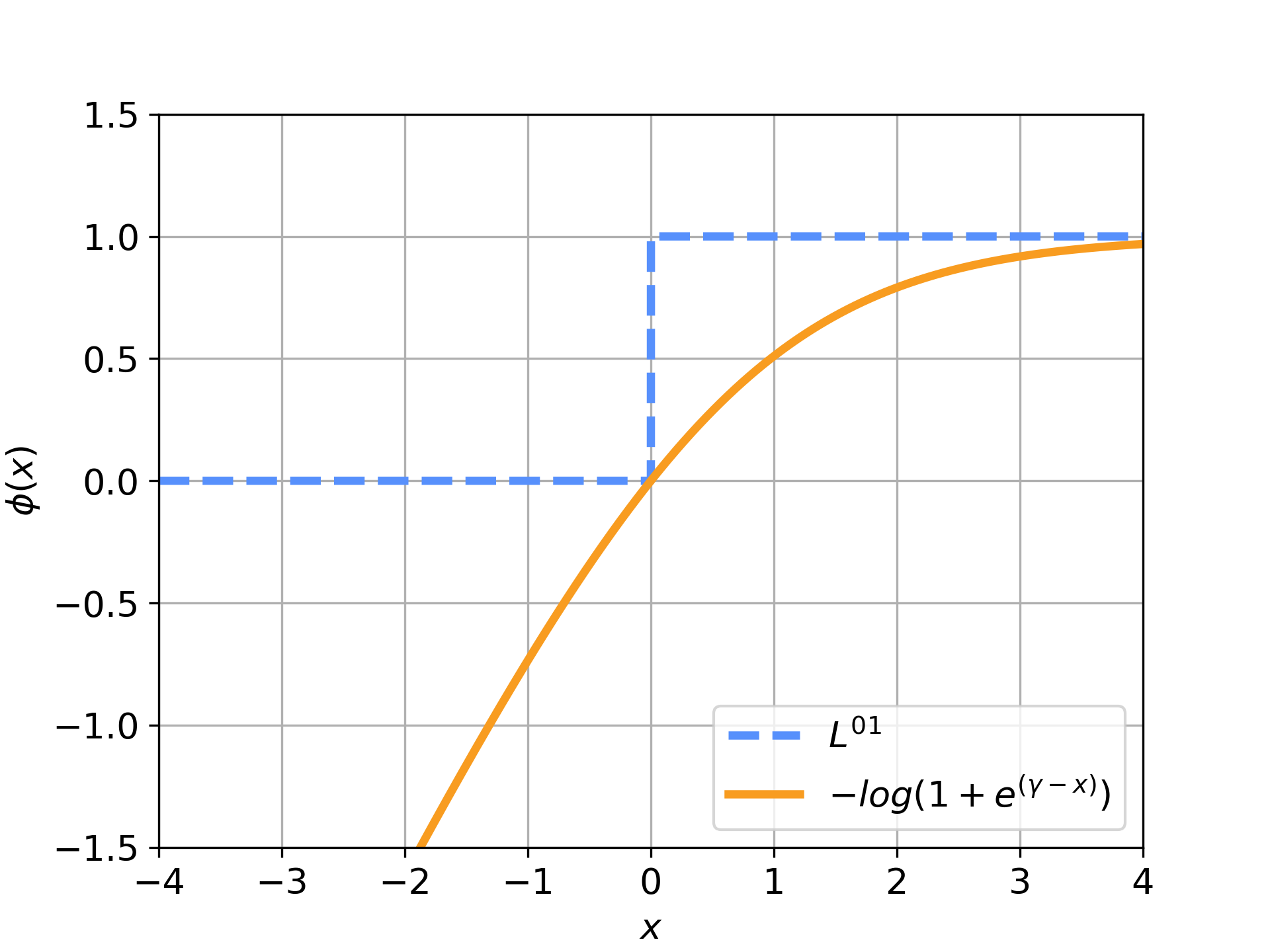}}
     \caption{ Plot of suggested lower bounds. }
     \label{fig:comparison_lower_bounds}
\end{figure}
\vspace{-1mm}

As shown in \cref{fig:comparison_lower_bounds}, the function 
\begin{equation}\label{eq:tanh}
\phi(x)=\tanh(\alpha x)
\end{equation}
seems to be a good lower bound of the 0-1 step function. However, at all points
far from zero, the gradient quickly vanishes, which might harm performance. We suggest considering the alternative:
\begin{equation}\label{eq:logloss}
\phi(x) = -\log(1 + e^{(\gamma-x)}) + 1
\end{equation}
where $\gamma=\log(e-1)$, also shown in \cref{fig:comparison_lower_bounds}, as it only suffers from a vanishing gradient on large positive values. However, note that
having a vanishing gradient in such a region is not really an issue: the objective of the perturbation $\epsilon$ is to move points towards the lower side of the plot in \cref{fig:comparison_lower_bounds}, where misclassification happens. Hence, if a point stays there due to the vanishing gradient problem, it means it will remain misclassified. In contrast, having vanishing gradients on the top side of the plot in \cref{fig:comparison_lower_bounds} might mean that the optimization algorithm is unable to move points that are correctly classified towards the misclassification region, therefore the adversary would fail.

\begin{figure}[!ht]
    \centering
    \includegraphics[width=0.4\textwidth]{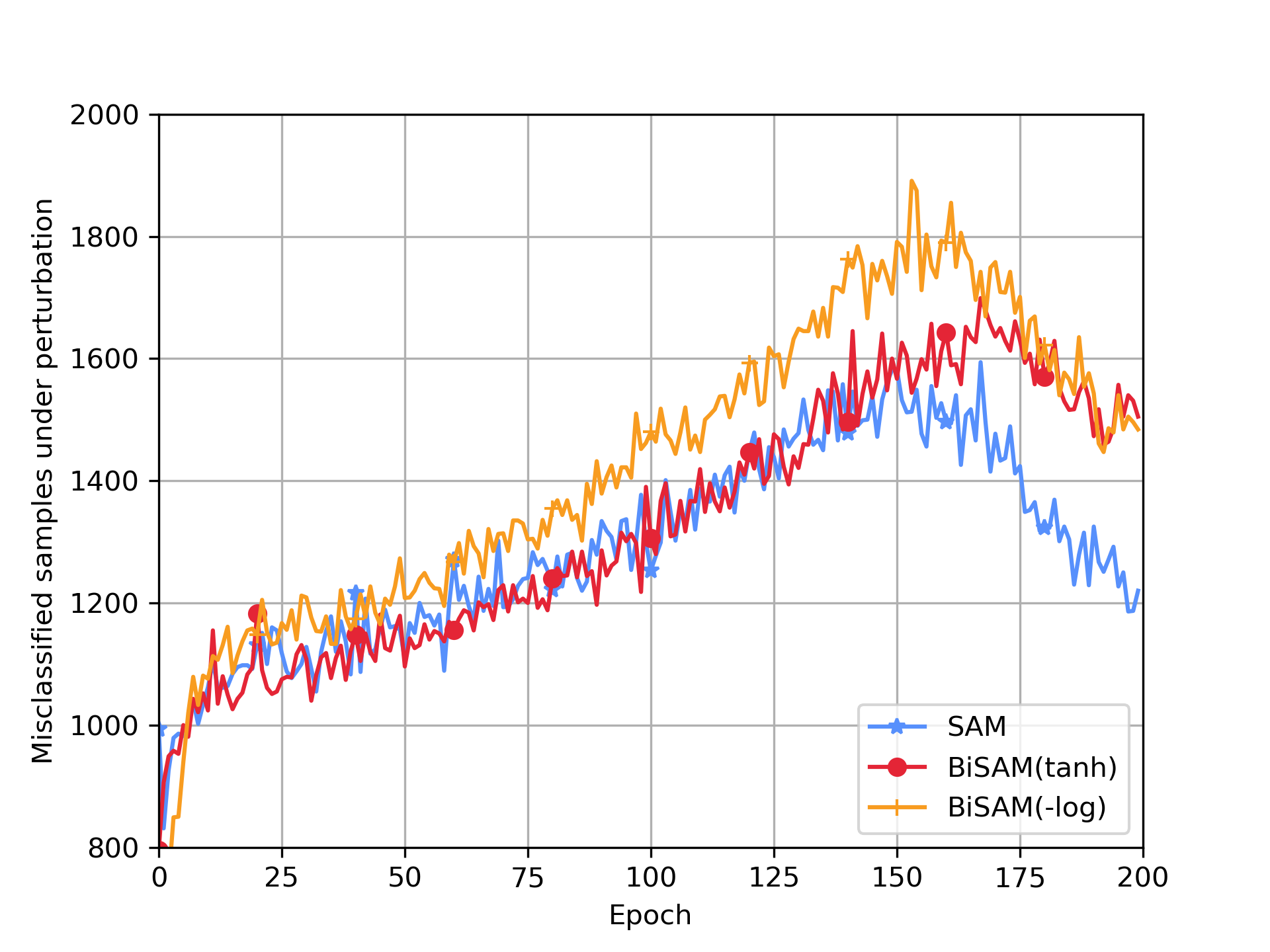}
    \caption{Number of misclassified samples under perturbation of Resnet56 on CIFAR-10.}
    \label{fig:reversed_data}
\vspace{-1mm}
\end{figure}

Both \cref{eq:tanh} and \cref{eq:logloss} used in $Q_{\phi,\mu}(w)$ serve as valid lower bounds in implementation.
\cref{fig:reversed_data} shows the number of misclassified samples under perturbation that the model predicts correctly while turning wrong after adding the weight perturbation. Note that its detailed setting is in \cref{sec:exp:classification}. It indicates that SAM indeed exhibits weaker perturbation compared to our proposed BiSAM. 

\vspace{-1mm}

\section{Experiments}
\label{sec:bisam_experiments}
In this section, we verify the benefit of BiSAM across a variety of models, datasets and tasks.

\begin{table*}[b]
    \centering
    \caption{\textbf{Test accuracies on CIFAR-10.} BiSAM (-log) has strictly better performance than SAM across \emph{all} models. 
    We include BiSAM (tanh) for completeness which sometimes performs better than BiSAM (-log).
    }
    \begin{tabular}{C{2.95cm} C{2.95cm} C{2.95cm} C{2.95cm} |>{\color{gray}} C{2.95cm}}
\toprule
\textbf{Model} & \textbf{SGD} & \textbf{SAM} & \textbf{BiSAM (-log)}  & \textbf{BiSAM (tanh)}  \\
\midrule                                                                 
       DenseNet-121     &   $96.14_{\pm 0.09}$    & 	$96.52_{\pm	0.10}$    &  $\mathbf{96.61}_{\pm 0.17}$ & $96.63_{\pm 0.21}$    \\
       Resnet-56        &   $94.01_{\pm 0.26}$    & 	$94.09_{\pm 0.26}$   &  $\mathbf{94.28}_{\pm 0.31}$ & $94.87_{\pm 0.34}$
						\\
       VGG19-BN       &     $94.76_{\pm 0.10}$  &  $95.09_{\pm	0.12}$   &		$\mathbf{95.22}_{\pm	0.13}$		& 	$95.01_{\pm	0.06}$	\\
       WRN-28-2      &      $95.71_{\pm 0.19}$     &	$96.00_{\pm 0.10}$ &		$\mathbf{96.02}_{\pm	0.12}$  	 &	 $95.99_{\pm 0.09}$		\\
       WRN-28-10  &     $96.77_{\pm 0.21}$     &   $97.18_{\pm	0.04}$    &		 $\mathbf{97.26}_{\pm 0.10}$    &  	$97.17_{\pm	0.05}$   \\
\midrule
    Average  & $95.48_{\pm 0.08}$  &  $95.78_{\pm 0.06}$    & $\mathbf{95.88}_{\pm 0.08}$  &  $95.93_{\pm 0.08}$ \\
\bottomrule
    \end{tabular}
    \label{tab:cifar10}
\end{table*}

\vspace{-1mm}
\begin{table*}[ht]
    \centering
    \caption{\textbf{Test accuracies on CIFAR-100.} BiSAM (-log) consistently improves over SAM across all models. 
    We include BiSAM (tanh) for completeness which sometimes performs better than BiSAM (-log).}
    \begin{tabular}{C{2.95cm} C{2.95cm} C{2.95cm} C{2.95cm} |>{\color{gray}} C{2.95cm}}
\toprule
\textbf{Model} & \textbf{SGD} & \textbf{SAM} & \textbf{BiSAM (-log)}  & \textbf{BiSAM (tanh)}  \\
\midrule                                                                 
       DenseNet-121     &    $81.31_{\pm 0.38}$         & 	$82.31_{\pm	0.15}$ &	$\mathbf{82.49}_{\pm 0.14}$ 		& 	$82.88_{\pm	0.42}$  	\\
       Resnet-56        &   $73.98_{\pm 0.16}$      & 	$74.38_{\pm 0.37}$ &		$\mathbf{74.67}_{\pm	0.15}$ 	 &	 $74.54_{\pm 0.35}$ 	\\
       VGG19-BN       &  $74.90_{\pm 0.30}$      &  $74.94_{\pm 0.12}$  &  $\mathbf{75.25}_{\pm	0.24}$      & 	$75.12_{\pm 0.34}$				\\
       WRN-28-2    &   $77.95_{\pm 0.14}$     &	$78.09_{\pm	0.13}$  &  $\mathbf{78.21}_{\pm 0.23}$ &	$78.07_{\pm	0.13}$ 				\\
       WRN-28-10  &    $81.50_{\pm	0.48}$    &   $82.89_{\pm 0.47}$ &	$\mathbf{83.27}_{\pm 0.26}$     &  $83.35_{\pm 0.25}$  		\\
\midrule
    Average  & $77.93_{\pm 0.14}$  &  $78.52_{\pm 0.13}$    & $\mathbf{78.78}_{\pm 0.09}$  &  $78.79_{\pm 0.14}$ \\
\bottomrule
    \end{tabular}
    \label{tab:cifar100}
    \vspace{-1mm}
\end{table*}

\subsection{Image classification}
\label{sec:exp:classification}

We follow the experimental setup of \citet{asam21}.
We use the CIFAR-10 and CIFAR-100 datasets \citep{krizhevsky2009learning}, both consisting of $50\,000$ training images of size $32\times 32$, with $10$ and $100$ classes, respectively. For data augmentation we apply the commonly used random cropping after padding with $4$ pixels, horizontal flipping, and normalization using the statistics of the training distribution at both train and test time. We train multiple variants of VGG \citep{simonyan2014very}, ResNet \citep{he2016deep}, DenseNet \citep{huang2017densely} and WideResNet \citep{zagoruyko2016wide} (see \Cref{tab:cifar10,tab:cifar100} for details) using cross entropy loss.
All experiments are conducted on an NVIDIA A100 GPU.

Two variants of BiSAM are compared against two baselines.
\begin{description}
\item[SGD:] Standard training using stochastic gradient descent (SGD) (see details below)
\item[SAM:] The original Sharpness-Aware Minimization (SAM) algorithm from \citet{sam20}
\item[BiSAM (tanh):] \Cref{alg:BiSAM} using \eqref{eq:tanh} as the lower bound
\item[BiSAM (-log):] \Cref{alg:BiSAM} using \eqref{eq:logloss} as the lower bound
\end{description}

The models are trained using stochastic gradient descent (SGD) with a momentum of $0.9$ and a weight decay of $5 \times 10^{-4}$. We used a batch size of $128$, and a cosine learning rate schedule that starts at $0.1$. The number of epochs is set to $200$ for SAM and BiSAM while SGD are given $400$ epochs. This is done in order to provide a computational fair comparison as (Bi)SAM uses twice as many gradient computation. Label smoothing with a factor $0.1$ is employed for all method.
For the SAM and BiSAM hyperparameter $\rho$ we use a value of $0.05$. 
We fix $\mu=10$ and $\alpha=0.1$ for BiSAM (tanh) and $\mu=1$ for BiSAM (-log) throughout \emph{all} experiments on both CIFAR-10 and CIFAR-100 datasets as a result of a grid search over $\{ 0.01, 0.1, 1, 10, 100 \}$ for $\alpha$ and over $\{0.1, 0.5, 1, 2, 4 \}$ for $\mu$ using the validation dataset on CIFAR-10 with Resnet-56.\looseness=-1

The training data is randomly partitioned into a training set and validation set consisting of $90\%$ and $10\%$, respectively.
We deviate from \citet{sam20,asam21} by using the \emph{validation} set to select the model on which we report the \emph{test} accuracy in order to avoid overfitting on the test set.
We report the test accuracy of the model with the highest validation accuracy across the training with mean and standard deviations computed over $6$ independent executions. The results can be found in \Cref{tab:cifar10,tab:cifar100}.
Compared to the accuracy increase from SGD to SAM, the average improvement from BiSAM to SAM reaches $33.3\%$ on CIFAR-10 and $44.1\%$ on CIFAR-100.

For evaluations at a larger scale, we compare the performance of SAM and BiSAM on ImageNet-1K~\citep{imagenet15}. We apply each method with $\rho=0.05$ for both SAM and BiSAM. We use training epochs $90$, peak learning rate $0.2$, and batch size $512$. We employ mSAM~\citep{sam20,msam23} with micro batch size $m=128$ to accelerate training and improve performance. We set $\mu=5$ for BiSAM (-log) and $\mu=20$ and $\alpha=0.1$ for BiSAM (tanh). Other parameters are the same as CIFAR-10 and CIFAR-100. We run $3$ independent experiments for each method and results are shown in \cref{tab:imagenet}. Note that we do not reproduce experiments of SGD on ImageNet-1K due to computational restriction but it well-documented that SAM and its variants have better performance than SGD \citep{sam20,asam21,du2022efficient}.

\begin{table}[!ht]
    \centering
    \caption{Test accuracies on ImageNet-1K.}
    \begin{tabular}{m{0.6cm}m{1.4cm}m{2cm} | >{\color{gray}}m{2.1cm}}
\toprule
\textbf{} & \textbf{SAM}  & \textbf{BiSAM (-log)} & \textbf{BiSAM (tanh)}  \\
\midrule                                                           
       Top1    &   $75.83_{\pm 0.16}$   & 	$\mathbf{75.96}_{\pm 0.15}$     & $76.02_{\pm 0.08}$ \\
       Top5    &   $92.47_{\pm 0.02}$ &	$\mathbf{92.49}_{\pm 0.10}$  & $92.40_{\pm 0.13}$       \\
\bottomrule
    \end{tabular}
    \label{tab:imagenet}
\end{table}

We find that BiSAM (-log) \emph{consistently} outperforms SGD and SAM across all models on both CIFAR-10 and CIFAR-100, and it also outperforms SAM on ImageNet-1K. In most cases, BiSAM (tanh) has better or almost same performance than SAM.
Average accuracies across $5$ models of both BiSAM (-log) and BiSAM (tanh) outperform SAM and the result is statistically significant as shown by the small standard deviation when aggregated over all model types.
This improvement is achieved \emph{without} modifying the original experimental setup and the hyperparameter involved.
Specifically, we use the same $\rho=0.05$ for BiSAM which has originally been tuned for SAM.
The consistent improvement using BiSAM, despite the favorable setting for SAM, shows the benefit of our reformulation based on the 0-1 loss.
Note that the generalization improvement provided by BiSAM comes at essentially \emph{no computational overhead} (see \cref{supp:computation} for detailed discussion).

We recommend using \textbf{BiSAM (-log)} as it generally achieves better or comparable test accuracies to BiSAM (tanh). Therefore, we choose BiSAM (-log) as representative while BiSAM (tanh) serves as reference in all tables.

\subsection{Incorporation with variants of SAM}
\label{sec:incorporation}
Since we just reformulate the perturbation loss of the original SAM, existing variants of SAM can be incorporated within BiSAM. The mSAM variant has been combined with BiSAM in experiments on ImageNet-1K. 
Moreover, we incorporate BiSAM with both Adaptive SAM \citep{asam21} and Efficient SAM \citep{du2022efficient}.

\paragraph{Adaptive BiSAM.} We combine BiSAM with Adaptive Sharpness in ASAM \citep{asam21} which proposes a normalization operator to realize adaptive sharpness. The Adaptive BiSAM (A-BiSAM) algorithm is specified in detail in \cref{supp:abisam} and results on CIFAR-10 are shown in \cref{tab:adaptive_cifar10}. A-BiSAM (-log) \emph{consistently} outperforms ASAM across \emph{all} models on CIFAR-10 except for one on DenseNet-121 where the accuracy is the same.

\vspace{-1mm}
\begin{table}[hb]
    \centering
    \caption{Test accuracies of A-(Bi)SAM.}
    \begin{tabular}{m{2.2cm}m{2.4cm}m{2.4cm}}
\toprule
\textbf{Model} & \textbf{ASAM}  & \textbf{A-BiSAM (-log)} \\
\midrule                                                           
      DenseNet-121    &   $\mathbf{96.79}_{\pm 0.14}$   & 	$\mathbf{96.79}_{\pm	0.13}$  \\
      Resnet-56    &   $94.86_{\pm 0.18}$ &	$\mathbf{95.09}_{\pm 0.09}$  \\
      VGG19-BN    &   $95.10_{\pm 0.09}$   & 	$\mathbf{95.14}_{\pm	0.14}$ \\
      WRN-28-2    &   $96.22_{\pm 0.10}$ &	$\mathbf{96.28}_{\pm 0.14}$   \\
      WRN-28-10    &   $97.37_{\pm 0.07}$   &  $\mathbf{97.42}_{\pm	0.09}$ \\
\midrule
    Average  & $96.07_{\pm 0.05}$  &  $\mathbf{96.14}_{\pm 0.05}$ \\
\bottomrule
    \end{tabular}
    \label{tab:adaptive_cifar10}
    \vspace{-1mm}
\end{table}

\paragraph{Efficient BiSAM.} %
BiSAM is also compatible with the two ideas constituting ESAM \citep{du2022efficient}, \textit{Stochastic Weight Perturbation} and \textit{Sharpness-sensitive Data Selection}. A detailed description of the combined algorithm, denoted E-BiSAM, is described in \cref{supp:ebisam} and results on CIFAR-10 are shown in \cref{tab:efficient_cifar10}. E-BiSAM (-log) improves the performance of ESAM across \emph{all} models on CIFAR-10.

\vspace{-1mm}
\begin{table}[ht]
    \centering
    \caption{Test accuracies of E-(Bi)SAM.}
    \begin{tabular}{m{2.2cm}m{2.4cm}m{2.4cm}}
\toprule
\textbf{Model} & \textbf{ESAM}  & \textbf{E-BiSAM (-log)} \\
\midrule                                                           
      DenseNet-121    &   $96.30_{\pm 0.22}$   & 	$\mathbf{96.35}_{\pm	0.12}$ \\
      Resnet-56    &   $94.21_{\pm 0.38}$ &	$\mathbf{94.60}_{\pm 0.24}$      \\
      VGG19-BN    &   $94.16_{\pm 0.09}$   & 	$\mathbf{94.43}_{\pm	0.14}$  \\
      WRN-28-2    &   $95.95_{\pm 0.08}$ &	$\mathbf{96.00}_{\pm 0.04}$       \\
      WRN-28-10    &   $97.17_{\pm 0.09}$   & 	$\mathbf{97.18}_{\pm	0.05}$ \\
\midrule
    Average  & $95.56_{\pm 0.09}$  &  $\mathbf{95.71}_{\pm 0.06}$ \\
\bottomrule
    \end{tabular}
    \label{tab:efficient_cifar10}
    \vspace{-1mm}
\end{table}

\subsection{Fine-tuning on image classification}
We conduct experiments of transfer learning on ViT architectures. In particular, we use pretrained ViT-B/16 checkpoint from Visual Transformers \citep{ViT} and finetune the model on Oxford-flowers \citep{flowers08} and Oxford-IITPets \citep{pets12} datasets. We use AdamW as base optimizer with no weight decay under a linear learning rate schedule and gradient clipping with global norm 1. We set peak learning rate to $1e-4$ and batch size to $512$, and run $500$ steps with a warmup step of $100$. Note that for Flowers dataset, we choose $\mu=4$ for BiSAM(-log) and $\mu=20$ for BiSAM(tanh); and for Pets dataset, set $\mu=6$ for BiSAM(-log) and $\mu=20$ for BiSAM(tanh). The results in the table indicate that BiSAM benefits transfer learning.

\begin{table}[!ht]
\vspace{-1mm}
    \centering
    \caption{Test accuracies for image fine-tuning. 
    }
    \begin{tabular}{m{1cm}m{1.4cm}m{2cm} | >{\color{gray}}m{2.1cm}}
\toprule
\textbf{} & \textbf{SAM}  & \textbf{BiSAM (-log)} & \textbf{BiSAM (tanh)}  \\
\midrule                                                           
       Flowers    &   $98.79_{\pm 0.07}$   & 	$\mathbf{98.93}_{\pm	0.15}$     & $98.87_{\pm 0.08}$ \\
       Pets    &   $93.66_{\pm 0.48}$ &	$\mathbf{94.15}_{\pm 0.24}$  & $93.81_{\pm 0.45}$
						\\
\bottomrule
    \end{tabular}
    \label{tab:finetune}
    \vspace{-1mm}
\end{table}

\begin{table*}[!b]
\vspace{-2mm}
    \centering
    \caption{Experimental results of BERT-base finetuned on GLUE.
    }
    \begin{tabular}{c|c|cccccccccc}
\toprule
\multirow{2}*{\textbf{Method}} & \multirow{2}*{\textbf{GLUE}} & \textbf{CoLA} & \textbf{SST-2} & \multicolumn{2}{c}{\textbf{MRPC}} & \multicolumn{2}{c}{\textbf{QQP}} & \textbf{MNLI} & \textbf{QNLI} & \textbf{RTE} & \textbf{WNLI}    \\
\cline{3-12}
~ ~& & \textit{Mcc.}  & \textit{Acc.} & \textit{Acc.} & \textit{F1.} & \textit{Acc.} & \textit{F1.}  & \textit{Acc.} & \textit{Acc.} & \textit{Acc.} & \textit{Acc.}     \\
\midrule                                                                 
       AdamW    &  $73.21$  &  $56.63$  & $91.64$ & $85.64$ & $89.94$ & $90.18$ & $86.81$ &  $82.59$  & $89.78$ & $62.38$ & $26.41$ \\
       -w SAM  &  $75.60$   & $58.78$  & $92.29$ & $86.49$ & $90.51$ & $90.62$ & $87.56$ & $83.96$ & $90.36$ & $60.65$	& $41.20$\\
       -w BiSAM   &  $\mathbf{77.04}$ & $\mathbf{58.94}$ & $\mathbf{92.54}$ & $\mathbf{86.91}$ & $\mathbf{90.72}$	 & $\mathbf{90.70}$ & $\mathbf{87.68}$ & $\mathbf{84.00}$ & $\mathbf{90.53}$ & $\mathbf{62.74}$ & $\mathbf{49.53}$\\
\bottomrule
    \end{tabular}
    \label{tab:glue}
    \vspace{-1mm}
\end{table*}

\vspace{-1mm}
\begin{table*}[!t]
    \centering
    \caption{Test accuracies of ResNet-32 models trained on CIFAR-10 with label noise.}
    \begin{tabular}{C{2.95cm} C{2.95cm} C{2.95cm} C{2.95cm} |>{\color{gray}} C{2.95cm}}
\toprule
\textbf{Noise rate} & \textbf{SGD} & \textbf{SAM} & \textbf{BiSAM (-log)} & \textbf{BiSAM (tanh)}  \\
\midrule                                                                 
       $0\%$    &    $94.76_{\pm 0.14}$    & 	$94.95_{\pm	0.13}$ 		&  $\mathbf{94.98}_{\pm	0.17}$ 		& 	$95.01_{\pm	0.08}$ 	\\
       $20\%$        &  $88.65_{\pm	1.75}$        & 	$92.57_{\pm	0.24}$ &		$\mathbf{92.59}_{\pm	0.11}$	 & $92.35_{\pm	0.29}$ 		\\
       $40\%$       &   $84.24_{\pm 0.25}$      &    $\mathbf{89.03}_{\pm	0.09}$    &	$88.71_{\pm	0.23}$  	&   $88.86_{\pm 0.18}$			\\
       $60\%$    &     $76.29_{\pm 0.25}$    &		$82.77_{\pm	0.29}$ & 	$\mathbf{82.91}_{\pm	0.46}$ 	 & 	$82.87_{\pm	0.71}$ 	\\
       $80\%$  &     $44.44_{\pm	1.20}$      &    $44.68_{\pm	4.01}$     & 	$\mathbf{50.00}_{\pm 1.96}$	  &  	$48.57_{\pm	0.64}$			\\
\bottomrule
    \end{tabular}
    \label{tab:label_noise}
    \vspace{-1mm}
\end{table*}

\subsection{NLP Fine-tuning}
To check if BiSAM can benefit the natural language processing (NLP) domain, we show empirical text classification results in this section.
In particular, we use BERT-base model and finetune it on the GLUE datasets \citep{wang2018glue}. 
Note that we do not include STS-B because it is not a classification task, instead it is a regression task.
We use AdamW as base optimizer under a linear learning rate schedule and gradient clipping with global norm 1. We set the peak learning rate to $2e-5$ and batch size to $32$, and run $3$ epochs with an exception for MRPC and WNLI which are tiny and where we used $5$ epochs. We use BiSAM (-log) with the same hyperparameter as on CIFAR datasets in this experiment.
Note that we set $\rho=0.05$ for all datasets except for CoLA with $\rho=0.01$ and RTE with $\rho=0.005$.
We report the results computed over 10 independent executions in the \cref{tab:glue}, which demonstrates that BiSAM also benefits in NLP domain.

\subsection{Noisy labels task}

\label{supp:noisy}

We test on a task outside the i.i.d. setting that the method was designed for.
Following \citet{sam20} we consider label noise, where a fraction of the labels in the training set are corrupted to another label sampled uniformly at random.
Apart from the label perturbation, the experimental setup is otherwise the same as in \Cref{sec:exp:classification}, except for adjusting $\rho=0.01$ for SAM and BiSAM when noise rate is $80\%$, as the original $\rho=0.05$ causes failure for both methods. We find that BiSAM enjoys similar robustness to label noise as SAM despite not being specifically designed for the setting.

Across $4$ tasks, $6$ datasets, $9$ models, and $2$ incorporations with SAM variants, our experiments validate BiSAM's improvement broadly. A \textit{consistent improvement} is observed in all experiments, which also implies a widespread cumulative impact.
In addition, it is crucial to emphasize that BiSAM has the same computational complexity as SAM, which has been validated in the implementation. The detailed discussion can be found in \cref{supp:computation}.

\section{Related Work}
\label{sec:bisam_related}
The min-max zero-sum optimization template has been used in recent years in multiple applications beyond SAM \citep{sam20} e.g., in Adversarial Training \citep{madry2018towards,latorre2023finding} or Generative Adversarial Networks (GANs) \citep{goodfellow2014}. 

In particular, the SAM formulation as a two-player game interacting via addition, has precedence in Robust Bayesian Optimization \citep{bogunovic2018adversarially}, where it is called $\epsilon$-perturbation stability. Even though our formulation starts as a zero-sum game \eqref{eq:sam01}, a tractable reformulation 
\eqref{eq:bilevel3} requires leveraging the bilevel optimization approach \citep{bard2013practical}. 

The bilevel paradigm has already seen applications in Machine Learning, with regard to hyperparameter optimization \citep{domke12,lorraine20a,mackay2018selftuning,Franceschi2018BilevelPF}, meta-learning \citep{Franceschi2018BilevelPF,rajeswaran2019meta}, data denoising by importance learning \citep{Ren2018LearningTR}, neural architecture search \citep{liu2018darts}, training data poisoning \citep{mei2015using,MuozGonzlez2017TowardsPO,huang2020metapoison}, and adversarial training \citep{robey2023adversarial}. Our formulation is the first bilevel formulation in the context of SAM. 

ESAM \citep{du2022efficient} introduces two tricks, \textit{Stochastic Weight Perturbation (SWP)} and \textit{Sharpness-sensitive Data Selection (SDS)} that subset random variables of the optimization problem, or a subset of the elements in the mini-batch drawn in a given iteration. Neither modification is related to the optimization objective of SAM. Thus, analogous ideas can be used inside our bilevel approach as shown in \cref{alg:EBiSAM}. This is useful, as ESAM can reduce the computational complexity of SAM while retaining its performance. We can see a similar result when combined with BiSAM. \looseness=-1

In ASAM \citep{asam21}, a notion of \textit{Adaptive Sharpness} is introduced, whereby the constraint set of the perturbation $\epsilon$ in \eqref{eq:sam} is modified to depend on the parameter $w$. This particular choice yields a definition of sharpness that is invariant under transformations that do not change the value of the loss. The arguments in favor of adaptive sharpness hold for arbitrary losses, and hence, adaptivity can also be incorporated within the bilevel formulation of BiSAM as shown in \cref{alg:ABiSAM}. Experimental results in \cref{tab:adaptive_cifar10} demonstrate that this incorporation improves performance.

A relationship between the inner-max objective in SAM and a Bayesian variational formulation was revealed by \citet{mollenhoff2023sam}. Based on this result, they proposed \textit{Bayesian SAM (bSAM)}, a modification of SAM that can obtain uncertainty estimates. While such results require a continuity condition on the loss c.f. \citet[Theorem 1.]{mollenhoff2023sam}, their arguments could be applied to any sufficiently tight continuous approximation of the 0-1 loss. Therefore, a similar relationship between our formulation of SAM and the Bayesian perspective could be derived, enabling uncertainty estimates for BiSAM. 

In GSAM \citep{zhuang2022surrogate}, propose to minimize the so-called \textit{surrogate gap} $\max_{\epsilon: \|\epsilon \| \leq \rho } L_S(w+\epsilon) - L_S(w)$ and the perturbed loss $\max_{\epsilon: \|\epsilon \| \leq \rho } L_S(w+\epsilon)$ simultaneously, which leads to a modified SAM update. In \citet{liu2022random}, it is proposed to add a random initialization before the optimization step that defines the perturbation. In \citet{ni2022ksam}, it is suggested that using the top-k elements of the mini-batch to compute the stochastic gradients is a good alternative to improve the speed of SAM. To different degrees, such SAM variants have analogous versions in our framework.\looseness=-1

\section{Conclusions and Future Work}
\label{sec:bisam_conclusions}

In this work, we proposed a novel formulation of SAM by utilizing the 0-1 loss for classification tasks. By reformulating SAM as a bilevel optimization problem, we aimed to maximize the lower bound of the 0-1 loss through perturbation. We proposed BiSAM, a scalable first-order optimization method, to effectively solve this bilevel optimization problem. 
Through experiments, BiSAM outperformed SAM in training, fine-tuning, and noisy label tasks meanwhile maintaining a similar computational complexity. 
In addition, incorporating variants of SAM (\eg ASAM, ESAM) in BiSAM can improve its performance or efficiency further. 

Given the promising results in classification tasks, exploring BiSAM's applicability in other domains such as text generation could broaden its scope.
In addition, we claim that the generalization bound, a motivation of SAM, remains valid for 0-1 loss as presented in \eqref{eq:sam01}. However, relating the solution of this to the solution of BiSAM shown in \eqref{eq:bilevel3} is still an open problem, which we leave for future work. We discuss this in detail in \cref{supp:generalization}.
Overall, the insights gained from this work offer new directions and opportunities for advancing the field of sharpness-aware optimization.

\section*{Acknowledgements}
We thank the reviewers for their constructive feedback. Thanks to Fanghui Liu for the helpful discussion.
This work was supported by the Swiss National Science Foundation (SNSF) under grant number 200021\_205011.
This work was funded through a PhD fellowship of the Swiss Data Science Center, a joint venture between EPFL and ETH Zurich.
This work was supported by Google.
This work was supported by Hasler Foundation Program: Hasler Responsible AI (project number 21043).
This research was sponsored by the Army Research Office and was accomplished under Grant Number W911NF-24-1-0048.

\section*{Impact Statement}
This paper presents work whose goal is to advance the field of Machine Learning. There are many potential societal consequences of our work, none of which we feel must be specifically highlighted here.

\bibliography{ref}

\begin{thebibliography}{42}
\providecommand{\natexlab}[1]{#1}
\providecommand{\url}[1]{\texttt{#1}}
\expandafter\ifx\csname urlstyle\endcsname\relax
  \providecommand{\doi}[1]{doi: #1}\else
  \providecommand{\doi}{doi: \begingroup \urlstyle{rm}\Url}\fi

\bibitem[Bahri et~al.(2021)Bahri, Mobahi, and Tay]{Bahri2021SharpnessAwareMI}
Dara Bahri, Hossein Mobahi, and Yi~Tay.
\newblock Sharpness-aware minimization improves language model generalization.
\newblock In \emph{Annual Meeting of the Association for Computational Linguistics}, 2021.

\bibitem[Bard(2013)]{bard2013practical}
Jonathan~F Bard.
\newblock \emph{Practical bilevel optimization: algorithms and applications}, volume~30.
\newblock 2013.

\bibitem[Behdin et~al.(2023)Behdin, Song, Gupta, Acharya, Durfee, Ocejo, Keerthi, and Mazumder]{msam23}
Kayhan Behdin, Qingquan Song, Aman Gupta, Ayan Acharya, David Durfee, Borja Ocejo, Sathiya Keerthi, and Rahul Mazumder.
\newblock msam: Micro-batch-averaged sharpness-aware minimization.
\newblock \emph{arXiv preprint arXiv:2302.09693}, 2023.

\bibitem[Bogunovic et~al.(2018)Bogunovic, Scarlett, Jegelka, and Cevher]{bogunovic2018adversarially}
Ilija Bogunovic, Jonathan Scarlett, Stefanie Jegelka, and Volkan Cevher.
\newblock Adversarially robust optimization with gaussian processes.
\newblock \emph{Advances in Neural Information Processing Systems (NeurIPS)}, 31, 2018.

\bibitem[Bolte(2020)]{benblog}
Ben Bolte.
\newblock Optimized log-sum-exp pytorch function.
\newblock \url{https://ben.bolte.cc/logsumexp}, 2020.

\bibitem[Domke(2012)]{domke12}
Justin Domke.
\newblock Generic methods for optimization-based modeling.
\newblock In \emph{International Conference on Artificial Intelligence and Statistics (AISTATS)}, 2012.

\bibitem[Dosovitskiy et~al.(2020)Dosovitskiy, Beyer, Kolesnikov, Weissenborn, Zhai, Unterthiner, Dehghani, Minderer, Heigold, Gelly, et~al.]{dosovitskiy2021an}
Alexey Dosovitskiy, Lucas Beyer, Alexander Kolesnikov, Dirk Weissenborn, Xiaohua Zhai, Thomas Unterthiner, Mostafa Dehghani, Matthias Minderer, Georg Heigold, Sylvain Gelly, et~al.
\newblock An image is worth 16x16 words: transformers for image recognition at scale.
\newblock In \emph{International Conference on Learning Representations (ICLR)}, 2020.

\bibitem[Du et~al.(2022)Du, Yan, Feng, Zhou, Zhen, Goh, and Tan]{du2022efficient}
Jiawei Du, Hanshu Yan, Jiashi Feng, Joey~Tianyi Zhou, Liangli Zhen, Rick Siow~Mong Goh, and Vincent Tan.
\newblock Efficient sharpness-aware minimization for improved training of neural networks.
\newblock In \emph{International Conference on Learning Representations (ICLR)}, 2022.

\bibitem[Dziugaite and Roy(2017)]{Dziugaite2017}
Gintare~Karolina Dziugaite and Daniel~M. Roy.
\newblock Computing nonvacuous generalization bounds for deep (stochastic) neural networks with many more parameters than training data.
\newblock In Gal Elidan, Kristian Kersting, and Alexander Ihler, editors, \emph{Uncertainty in Artificial Intelligence Conference}, 2017.

\bibitem[Foret et~al.(2021)Foret, Kleiner, Mobahi, and Neyshabur]{sam20}
Pierre Foret, Ariel Kleiner, Hossein Mobahi, and Behnam Neyshabur.
\newblock Sharpness-aware minimization for efficiently improving generalization.
\newblock In \emph{International Conference on Learning Representations (ICLR)}, 2021.

\bibitem[Franceschi et~al.(2018)Franceschi, Frasconi, Salzo, Grazzi, and Pontil]{Franceschi2018BilevelPF}
Luca Franceschi, Paolo Frasconi, Saverio Salzo, Riccardo Grazzi, and Massimiliano Pontil.
\newblock Bilevel programming for hyperparameter optimization and meta-learning.
\newblock In \emph{International Conference on Machine Learning (ICML)}, 2018.

\bibitem[Ghadimi and Wang(2018)]{ghadimi2018approximation}
Saeed Ghadimi and Mengdi Wang.
\newblock Approximation methods for bilevel programming.
\newblock \emph{arXiv preprint arXiv:1802.02246}, 2018.

\bibitem[Goodfellow et~al.(2014)Goodfellow, Pouget-Abadie, Mirza, Xu, Warde-Farley, Ozair, Courville, and Bengio]{goodfellow2014}
Ian Goodfellow, Jean Pouget-Abadie, Mehdi Mirza, Bing Xu, David Warde-Farley, Sherjil Ozair, Aaron Courville, and Yoshua Bengio.
\newblock Generative adversarial nets.
\newblock \emph{Advances in Neural Information Processing Systems (NeurIPS)}, 27, 2014.

\bibitem[He et~al.(2016)He, Zhang, Ren, and Sun]{he2016deep}
Kaiming He, Xiangyu Zhang, Shaoqing Ren, and Jian Sun.
\newblock Deep residual learning for image recognition.
\newblock In \emph{Conference on Computer Vision and Pattern Recognition (CVPR)}, 2016.

\bibitem[Hochreiter et~al.(2001)Hochreiter, Bengio, Frasconi, et~al.]{hochreiter2001gradient}
Sepp Hochreiter, Yoshua Bengio, Paolo Frasconi, et~al.
\newblock Gradient flow in recurrent nets: the difficulty of learning long-term dependencies, 2001.

\bibitem[Huang et~al.(2017)Huang, Liu, Van Der~Maaten, and Weinberger]{huang2017densely}
Gao Huang, Zhuang Liu, Laurens Van Der~Maaten, and Kilian~Q Weinberger.
\newblock Densely connected convolutional networks.
\newblock In \emph{Conference on Computer Vision and Pattern Recognition (CVPR)}, 2017.

\bibitem[Huang et~al.(2020)Huang, Geiping, Fowl, Taylor, and Goldstein]{huang2020metapoison}
W~Ronny Huang, Jonas Geiping, Liam Fowl, Gavin Taylor, and Tom Goldstein.
\newblock Metapoison: Practical general-purpose clean-label data poisoning.
\newblock \emph{Advances in Neural Information Processing Systems (NeurIPS)}, 33, 2020.

\bibitem[Krizhevsky et~al.(2009)Krizhevsky, Hinton, et~al.]{krizhevsky2009learning}
Alex Krizhevsky, Geoffrey Hinton, et~al.
\newblock Learning multiple layers of features from tiny images.
\newblock 2009.

\bibitem[Kwon et~al.(2021)Kwon, Kim, Park, and Choi]{asam21}
Jungmin Kwon, Jeongseop Kim, Hyunseo Park, and In~Kwon Choi.
\newblock {ASAM}: Adaptive sharpness-aware minimization for scale-invariant learning of deep neural networks.
\newblock In \emph{International Conference on Machine Learning (ICML)}, 2021.

\bibitem[Latorre et~al.(2023)Latorre, Krawczuk, Dadi, Pethick, and Cevher]{latorre2023finding}
Fabian Latorre, Igor Krawczuk, Leello~Tadesse Dadi, Thomas~Michaelsen Pethick, and Volkan Cevher.
\newblock Finding actual descent directions for adversarial training.
\newblock In \emph{International Conference on Learning Representations (ICLR)}, 2023.

\bibitem[Liu et~al.(2018)Liu, Simonyan, and Yang]{liu2018darts}
Hanxiao Liu, Karen Simonyan, and Yiming Yang.
\newblock {DARTS}: Differentiable architecture search.
\newblock In \emph{International Conference on Learning Representations (ICLR)}, 2018.

\bibitem[Liu et~al.(2022)Liu, Mai, Cheng, Chen, Hsieh, and You]{liu2022random}
Yong Liu, Siqi Mai, Minhao Cheng, Xiangning Chen, Cho-Jui Hsieh, and Yang You.
\newblock Random sharpness-aware minimization.
\newblock \emph{Advances in Neural Information Processing Systems (NeurIPS)}, 2022.

\bibitem[Lorraine et~al.(2020)Lorraine, Vicol, and Duvenaud]{lorraine20a}
Jonathan Lorraine, Paul Vicol, and David Duvenaud.
\newblock Optimizing millions of hyperparameters by implicit differentiation.
\newblock In \emph{International Conference on Artificial Intelligence and Statistics (AISTATS)}, 2020.

\bibitem[Mackay et~al.(2019)Mackay, Vicol, Lorraine, Duvenaud, and Grosse]{mackay2018selftuning}
Matthew Mackay, Paul Vicol, Jonathan Lorraine, David Duvenaud, and Roger Grosse.
\newblock Self-tuning networks: Bilevel optimization of hyperparameters using structured best-response functions.
\newblock In \emph{International Conference on Learning Representations (ICLR)}, 2019.

\bibitem[Madry et~al.(2018)Madry, Makelov, Schmidt, Tsipras, and Vladu]{madry2018towards}
Aleksander Madry, Aleksandar Makelov, Ludwig Schmidt, Dimitris Tsipras, and Adrian Vladu.
\newblock Towards deep learning models resistant to adversarial attacks.
\newblock In \emph{International Conference on Learning Representations (ICLR)}, 2018.

\bibitem[Mei and Zhu(2015)]{mei2015using}
Shike Mei and Xiaojin Zhu.
\newblock Using machine teaching to identify optimal training-set attacks on machine learners.
\newblock In \emph{AAAI Conference on Artificial Intelligence}, volume~29, 2015.

\bibitem[M{\"o}llenhoff and Khan(2022)]{mollenhoff2023sam}
Thomas M{\"o}llenhoff and Mohammad~Emtiyaz Khan.
\newblock {SAM} as an optimal relaxation of {Bayes}.
\newblock In \emph{International Conference on Learning Representations (ICLR)}, 2022.

\bibitem[Mu{\~n}oz-Gonz{\'a}lez et~al.(2017)Mu{\~n}oz-Gonz{\'a}lez, Biggio, Demontis, Paudice, Wongrassamee, Lupu, and Roli]{MuozGonzlez2017TowardsPO}
Luis Mu{\~n}oz-Gonz{\'a}lez, Battista Biggio, Ambra Demontis, Andrea Paudice, Vasin Wongrassamee, Emil~C. Lupu, and Fabio Roli.
\newblock Towards poisoning of deep learning algorithms with back-gradient optimization.
\newblock \emph{Proceedings of the 10th ACM Workshop on Artificial Intelligence and Security}, 2017.

\bibitem[Ni et~al.(2022)Ni, Chiang, Geiping, Goldblum, Wilson, and Goldstein]{ni2022ksam}
Renkun Ni, Ping-yeh Chiang, Jonas Geiping, Micah Goldblum, Andrew~Gordon Wilson, and Tom Goldstein.
\newblock K-sam: Sharpness-aware minimization at the seed of {SGD}.
\newblock \emph{arXiv preprint arXiv:2210.12864}, 2022.

\bibitem[Nilsback and Zisserman(2008)]{flowers08}
Maria-Elena Nilsback and Andrew Zisserman.
\newblock Automated flower classification over a large number of classes.
\newblock In \emph{2008 Sixth Indian conference on computer vision, graphics \& image processing}, 2008.

\bibitem[Parkhi et~al.(2012)Parkhi, Vedaldi, Zisserman, and Jawahar]{pets12}
Omkar~M Parkhi, Andrea Vedaldi, Andrew Zisserman, and CV~Jawahar.
\newblock Cats and dogs.
\newblock In \emph{Conference on Computer Vision and Pattern Recognition (CVPR)}, 2012.

\bibitem[Rajeswaran et~al.(2019)Rajeswaran, Finn, Kakade, and Levine]{rajeswaran2019meta}
Aravind Rajeswaran, Chelsea Finn, Sham~M Kakade, and Sergey Levine.
\newblock Meta-learning with implicit gradients.
\newblock \emph{Advances in Neural Information Processing Systems (NeurIPS)}, 32, 2019.

\bibitem[Ren et~al.(2018)Ren, Zeng, Yang, and Urtasun]{Ren2018LearningTR}
Mengye Ren, Wenyuan Zeng, Binh Yang, and Raquel Urtasun.
\newblock Learning to reweight examples for robust deep learning.
\newblock In \emph{International Conference on Machine Learning (ICML)}, 2018.

\bibitem[Robey et~al.(2023)Robey, Latorre, Pappas, Hassani, and Cevher]{robey2023adversarial}
Alexander Robey, Fabian Latorre, George~J Pappas, Hamed Hassani, and Volkan Cevher.
\newblock Adversarial training should be cast as a non-zero-sum game.
\newblock \emph{arXiv preprint arXiv:2306.11035}, 2023.

\bibitem[Russakovsky et~al.(2015)Russakovsky, Deng, Su, Krause, Satheesh, Ma, Huang, Karpathy, Khosla, Bernstein, et~al.]{imagenet15}
Olga Russakovsky, Jia Deng, Hao Su, Jonathan Krause, Sanjeev Satheesh, Sean Ma, Zhiheng Huang, Andrej Karpathy, Aditya Khosla, Michael Bernstein, et~al.
\newblock Imagenet large scale visual recognition challenge.
\newblock \emph{International Journal of Computer Vision (IJCV)}, 115, 2015.

\bibitem[Simonyan and Zisserman(2014)]{simonyan2014very}
Karen Simonyan and Andrew Zisserman.
\newblock Very deep convolutional networks for large-scale image recognition.
\newblock \emph{arXiv preprint arXiv:1409.1556}, 2014.

\bibitem[Tarzanagh and Balzano(2022)]{tarzanagh2023online}
Davoud~Ataee Tarzanagh and Laura Balzano.
\newblock Online bilevel optimization: Regret analysis of online alternating gradient methods.
\newblock \emph{arXiv preprint arXiv:2207.02829}, 2022.

\bibitem[Wang et~al.(2018)Wang, Singh, Michael, Hill, Levy, and Bowman]{wang2018glue}
Alex Wang, Amanpreet Singh, Julian Michael, Felix Hill, Omer Levy, and Samuel~R Bowman.
\newblock {GLUE}: A multi-task benchmark and analysis platform for natural language understanding.
\newblock \emph{arXiv preprint arXiv:1804.07461}, 2018.

\bibitem[Wu et~al.(2020)Wu, Xu, Dai, Wan, Zhang, Yan, Tomizuka, Gonzalez, Keutzer, and Vajda]{ViT}
Bichen Wu, Chenfeng Xu, Xiaoliang Dai, Alvin Wan, Peizhao Zhang, Zhicheng Yan, Masayoshi Tomizuka, Joseph Gonzalez, Kurt Keutzer, and Peter Vajda.
\newblock Visual transformers: Token-based image representation and processing for computer vision.
\newblock \emph{arXiv preprint arXiv:2006.03677}, 2020.

\bibitem[Zagoruyko and Komodakis(2016)]{zagoruyko2016wide}
Sergey Zagoruyko and Nikos Komodakis.
\newblock Wide residual networks.
\newblock \emph{arXiv preprint arXiv:1605.07146}, 2016.

\bibitem[Zhong et~al.(2022)Zhong, Ding, Shen, Mi, Liu, Du, and Tao]{sam-language22}
Qihuang Zhong, Liang Ding, Li~Shen, Peng Mi, Juhua Liu, Bo~Du, and Dacheng Tao.
\newblock Improving sharpness-aware minimization with fisher mask for better generalization on language models.
\newblock \emph{arXiv preprint arXiv:2210.05497}, 2022.

\bibitem[Zhuang et~al.(2021)Zhuang, Gong, Yuan, Cui, Adam, Dvornek, s~Duncan, Liu, et~al.]{zhuang2022surrogate}
Juntang Zhuang, Boqing Gong, Liangzhe Yuan, Yin Cui, Hartwig Adam, Nicha~C Dvornek, James s~Duncan, Ting Liu, et~al.
\newblock Surrogate gap minimization improves sharpness-aware training.
\newblock In \emph{International Conference on Learning Representations (ICLR)}, 2021.

\end{thebibliography}

\clearpage
\onecolumn
\appendix

\section{Complete experiments}
\label{supp:experiments}
We provide complete algorithms and experimental results here as complementary for \cref{sec:incorporation}.

\subsection{Adaptive BiSAM}
\label{supp:abisam}
As we introduced in \cref{sec:bisam_related}, some existing variants of SAM can be incorporated within BiSAM. To demonstrate this, we combine BiSAM with Adaptive Sharpness in ASAM \citep{asam21}. \citet{asam21} propose that the fixed radius of SAM's neighborhoods has a weak correlation with the generalization gap. Therefore, ASAM proposes a normalization operator to realize adaptive sharpness. Following the element-wise operator in \citet{asam21} defined by
\begin{equation}
    T_w = \diag(|w_1|, \ldots ,|w_b|),  \text{ where } w=[w_1, \ldots, w_b].
\end{equation}
We construct Adaptive BiSAM (A-BiSAM) in \cref{alg:ABiSAM}.
\vspace{-2mm}
\begin{figure}[!ht]
    \centering
\begin{minipage}{0.95\textwidth}
\begin{algorithm}[H]
\caption{Adaptive BiSAM (A-BiSAM)}
\label{alg:ABiSAM}
\SetAlgoLined
\setcounter{AlgoLine}{0}
    \KwIn{Initialization $w_0 \in \reals^d$, iterations $T$, batch size $b$, step sizes $\{\eta_t\}_{t=0}^{T-1}$, neighborhood size $\rho>0$, lower bound $\phi$.}
    \For{$t=0$ {\bfseries to} $T-1$}{
        Sample minibatch $\Bc=\{(x_1, y_1), \ldots, (x_b, y_b)\}$. \par
        Compute the (stochastic) gradient of the perturbation loss $Q_{\phi, \mu}(w_t)$ defined in \cref{eq:q_definition} \par
        Compute perturbation $\epsilon_t = \rho \frac{T_w^2 \nabla_w Q(w)}{\|T_w \nabla_w Q(w)\|}$ . \par
        Compute gradient $g_{t} = \nabla_w L_\Bc(w_t + \epsilon_t)$. \par
        Update weights $w_{t+1} = w_t - \eta_t g_{t}$.
    }
\end{algorithm}
\end{minipage}
\end{figure}

To compare A-BiSAM with ASAM, we use the same experimental setting as in \cref{sec:exp:classification} except for the choice of $\rho$. For both ASAM and A-BiSAM, we use $\rho=2$ as a result of a grid search over $\{0.1, 0.5, 1, 2, 3 \}$ using the validation dataset on CIFAR-10 with Resnet-56. (Note that we do not use $\rho=0.5$ as in \citet{asam21} because the results of ASAM with $\rho=0.5$ cannot outperform SAM in our experiments.)
We also have two variants of A-BiSAM compared against ASAM:
\begin{description}
\item[ASAM:] The original Adaptive Sharpness-Aware Minimization (ASAM) algorithm from \citet{asam21}
\item[A-BiSAM (tanh):] \Cref{alg:ABiSAM} using \eqref{eq:tanh} as lower bound
\item[A-BiSAM (-log):] \Cref{alg:ABiSAM} using \eqref{eq:logloss} as lower bound
\end{description}

We report the test accuracy of the model with the highest validation accuracy across the training with mean and standard deviations computed over $5$ independent executions.
The results can be found in  \Cref{tab:adaptive_cifar10_complete}.

\begin{table*}[hb]
    \centering
    \caption{Test accuracies of A-(Bi)SAM on CIFAR-10 dataset.}
    \begin{tabular}{C{3cm} C{3cm} C{3cm} C{3cm}}
\toprule
\textbf{Model} & \textbf{ASAM}  & \textbf{A-BiSAM (-log)}  & \textbf{A-BiSAM (tanh)} \\
\midrule                                                           
       DenseNet-121    &   $96.79_{\pm 0.14}$   & 	$96.79_{\pm	0.13}$  & 	$96.76_{\pm	0.06}$ \\
       Resnet-56    &   $94.86_{\pm 0.18}$ &	$95.09_{\pm 0.09}$ & 	$94.86_{\pm	0.12}$        \\
       VGG19-BN    &   $95.10_{\pm 0.09}$   & 	$95.14_{\pm	0.14}$ & 	$95.19_{\pm	0.15}$ \\
       WRN-28-2    &   $96.22_{\pm 0.10}$ &	$96.28_{\pm 0.14}$ & 	$96.29_{\pm	0.18}$       \\
       WRN-28-10    &   $97.37_{\pm 0.07}$   & 	$97.42_{\pm	0.09}$  & 	$97.34_{\pm	0.11}$   \\
\bottomrule
    \end{tabular}
    \label{tab:adaptive_cifar10_complete}
\end{table*}

\subsection{Efficient BiSAM}
\label{supp:ebisam}
A-BiSAM above mainly improves the performance of BiSAM while some variants of SAM can enhance the efficiency like Efficient SAM (ESAM) \citep{du2022efficient}. As we introduced in \cref{sec:bisam_related}, ESAM proposes two tricks, \textit{Stochastic Weight Perturbation (SWP)} and \textit{Sharpness-sensitive Data Selection (SDS)}, which can also be used in BiSAM.
When these two tricks are combined with BiSAM we refer to it as Efficient BiSAM (E-BiSAM) in \cref{alg:EBiSAM}.

\begin{figure}
    \centering
\begin{minipage}{0.95\textwidth}
\begin{algorithm}[H]
\caption{Efficient BiSAM (E-BiSAM)}
\label{alg:EBiSAM}
\SetAlgoLined
\setcounter{AlgoLine}{0}
    \KwIn{Initialization $w_0 \in \reals^d$, iterations $T$, batch size $b$, step sizes $\{\eta_t\}_{t=0}^{T-1}$, neighborhood size $\rho>0$, $\mu>0$, lower bound $\phi$, SWP hyperparameter $\beta$, SDS hyperparameter $\gamma$.}
    \For{$t=0$ {\bfseries to} $T-1$}{
        Sample minibatch $\Bc=\{(x_1, y_1), \ldots, (x_b, y_b)\}$. \par
        \For{$i=0$ {\bfseries to} $d-1$}{
            \eIf{$w_t[i]$ is chosen by probability $\beta$}{$\epsilon_t[i] = \frac{\rho}{1-\beta} \nabla_{w[i]} Q(w_t)$}{$\epsilon_t[i] = 0$}
        }
        Compute the perturbation loss $Q(w_t + \epsilon_t)$ and construct $\Bc^+$ with selection ratio $\gamma$:
        \begin{equation}
            \Bc^+ = \{ (x_i, y_i) \in \Bc: Q(w_t + \epsilon_t) - Q(w_t) > a\},  \text{ where $a$ controls } \gamma=\frac{|\Bc^+|}{|\Bc|}.
        \nn
        \end{equation} \par
        Compute gradient $g_{t} = \nabla_w L_{\Bc^+}(w_t + \epsilon_t)$. \par
        Update weights $w_{t+1} = w_t - \eta_t g_{t}$.
    }
\end{algorithm}
\end{minipage}
\end{figure}

To compare E-BiSAM with ESAM, we use the same experimental setting as in \cref{sec:exp:classification}. For hyperparameter $\beta$ and $\gamma$ for SWP and SDS respectively, we choose $0.5$ for both which is same as \citet{du2022efficient}.
We compare two variants of E-BiSAM against ESAM:
\begin{description}
\item[ESAM:] The original Efficient Sharpness-Aware Minimization (ESAM) algorithm from \citet{du2022efficient}
\item[E-BiSAM (tanh):] \Cref{alg:EBiSAM} using \eqref{eq:tanh} as lower bound
\item[E-BiSAM (-log):] \Cref{alg:EBiSAM} using \eqref{eq:logloss} as lower bound
\end{description}

We report the test accuracy of the model with the highest validation accuracy across the training with mean and standard deviations computed over $5$ independent executions.
The results can be found in  \Cref{tab:efficient_cifar10_complete}.

We observe that E-BiSAM (-log) outperformances ESAM across \emph{all} models and E-BiSAM (tanh) has same or better performance except for on WRN-18-10.
As a result, E-BiSAM combined with SWP and SDS improves the efficiency of BiSAM meanwhile keeping good performance.

\begin{table*}[ht]
    \centering
    \caption{Test accuracies of E-(Bi)SAM on CIFAR-10 dataset.}
    \begin{tabular}{C{3cm} C{3cm} C{3cm} C{3cm}}
\toprule
\textbf{Model} & \textbf{ESAM}  & \textbf{E-BiSAM (-log)}  & \textbf{E-BiSAM (tanh)} \\
\midrule                                                           
       DenseNet-121    &   $96.30_{\pm 0.22}$   & 	$96.35_{\pm	0.12}$  & 	$96.32_{\pm	0.11}$ \\
       Resnet-56    &   $94.21_{\pm 0.38}$ &	$94.60_{\pm 0.24}$ & 	$94.32_{\pm	0.34}$     \\
       VGG19-BN    &   $94.16_{\pm 0.09}$   & 	$94.43_{\pm	0.14}$ & 	$94.31_{\pm	0.12}$ \\
       WRN-28-2    &   $95.95_{\pm 0.08}$ &	$96.00_{\pm 0.04}$ & 	$95.95_{\pm	0.09}$  \\
       WRN-28-10    &   $97.17_{\pm 0.09}$   & 	$97.18_{\pm	0.05}$  & 	$97.14_{\pm	0.07}$ \\
\bottomrule
    \end{tabular}
    \label{tab:efficient_cifar10_complete}
\end{table*}

\section{Computational complexity}
\label{supp:computation}
We claim that BiSAM has the same computational complexity as SAM. 
This can be seen from the fact that the only change in BiSAM is the loss function used for the ascent step. By visual inspection of such loss function, its forward pass has the same complexity as that of vanilla SAM: we use the same logits but change the final loss function. Hence, the complexity should remain the same. We report timings of each epoch on CIFAR10 in our experiments. Note that time of training on CIFAR10 and CIFAR100 are roughly same. \looseness=-1

\begin{table}[!ht]
    \centering
    \caption{Time of each epoch.}
    \begin{tabular}{C{2.3cm} C{2.3cm} C{2.3cm} }
\toprule
\textbf{Model} & \textbf{SAM (cross-entropy)} & \textbf{BiSAM (logsumexp)}  \\
\midrule                                                                 
       DenseNet-121     & 58s  &  64s    \\
       Resnet-56   &  23s  &  30s     \\
       VGG19-BN      &  10s  &  16s   \\
       WRN-28-2      &  19s &  21s    \\
       WRN-28-10     & 65s  & 71s   \\
\bottomrule
    \end{tabular}
    \label{tab:time_cifar10}
\end{table}

The relatively small computational overhead ($10\%$ in the best cases) is most likely due to cross entropy being heavily optimized in Pytorch. There is no apparent reason why logsumexp should be slower so we expect that the gap can be made to effectively disappear if logsumexp is given similar attention. In fact, it has been pointed out before that logsumexp in particular is not well-optimized in Pytorch \citep{benblog}.

To provide further evidence that the computation overhead can be removed, we time the forward/backward of the ascent loss in both Pytorch and TensorFlow with batch size=128 and number of class=100 for 10k repetitions. We find that in tensorflow BiSAM would even enjoy a speedup over SAM.
\begin{table}[!hb]
    \centering
    \caption{Compare Pytorch with TensorFlow.}
    \begin{tabular}{C{2.3cm} C{2.3cm} C{2.3cm} }
\toprule
\textbf{Model} & \textbf{SAM (cross-entropy)} & \textbf{BiSAM (logsumexp)}  \\
\midrule                                       
      Pytorch	& 2.40s	& 3.96s    \\
      Tensorflow	& 3.25s &	2.34s    \\
\bottomrule
    \end{tabular}
    \label{tab:pytorch_tensorflow}
\end{table}

\section{Discussion}
\label{supp:generalization}

In this section we highlight that the generalization bounds provided in \citet{sam20} also holds when the loss is the (discontinuous) 0-1 loss.
We restate for convinience the theorem, which uses the PAC-Bayesian generalization bound of \citet{Dziugaite2017}.
\begin{theorem}{\citep[(stated informally)]{sam20}}\label{thm:sam_main_thm}
For any $\rho > 0$, with high probability over training set $S$ generated
from distribution $\mathcal{D}$:
\begin{equation}\label{eq:sam_main_eq}
L_{\mathcal{D}}(w) \leq \max_{\epsilon: \|\epsilon\|_2 \leq \rho} L_S(w+\epsilon) + h \left ( \|w\|^2_2 / \rho^2 \right)
\end{equation}
where $h: \mathbb{R}_+ \to \mathbb{R}_+$ is a strictly increasing function (under some technical conditions on $L_\mathcal{D}(w)$).
\end{theorem}

\begin{remark} 
Due to no specific differential assumptions in the proof of \cref{thm:sam_main_thm}, the bound also applies to the 0-1 loss, i.e.,
\begin{equation}
L^{01}_{\mathcal{D}}(w) \leq \max_{\epsilon: \|\epsilon\|_2 \leq \rho} L^{01}_S(w+\epsilon) + h \left ( \|w\|^2_2 / \rho^2 \right).
\end{equation}
This generalization bound provides motivation for solving the minimax problem over the 0-1 loss as given in \eqref{eq:sam01}.
It remains open to relate the solution of the bilevel optimization relaxation \eqref{eq:bilevel3} to a solution of \eqref{eq:sam01}.
\end{remark}

\end{document}